\def\year{2022}\relax
\documentclass[letterpaper]{article} 
\usepackage{aaai22}  
\usepackage{times}  
\usepackage{helvet}  
\usepackage{courier}  
\usepackage[hyphens]{url}  
\usepackage{graphicx} 
\usepackage{natbib}  
\usepackage{caption} 
\DeclareCaptionStyle{ruled}{labelfont=normalfont,labelsep=colon,strut=off} 
\usepackage{algorithm}
\usepackage{algorithmic}
\usepackage{comment}
\usepackage{color}
\usepackage{amsmath}
\usepackage{amsfonts}
\usepackage[switch]{lineno} 

\usepackage{newfloat}
\usepackage{listings}
\floatstyle{ruled}
\newfloat{listing}{tb}{lst}{}
\floatname{listing}{Listing}
\title{MobileFaceSwap: A Lightweight Framework for Video Face Swapping}

\author{
Zhiliang Xu \textsuperscript{\rm 1}, 
Zhibin Hong \textsuperscript{\rm 1} \thanks{Corresponding author.}, 
Changxing Ding \textsuperscript{\rm 2}, 
Zhen Zhu \textsuperscript{\rm 3}, \\
Junyu Han \textsuperscript{\rm 1}, 
Jingtuo Liu \textsuperscript{\rm 1}, 
Errui Ding \textsuperscript{\rm 1}
}
\affiliations{
    \textsuperscript{\rm 1} Baidu Inc. \\
    \textsuperscript{\rm 2} South China University of Technology \\
    \textsuperscript{\rm 3} University of Illinois at Urbana-Champaign \\
    \{xuzhiliang, hongzhibin, hanjunyu, liujintuo, dingerrui\}@baidu.com, \\ chxding@scut.edu.cn, zhenzhu4@illinois.edu

}

\begin{document}

\maketitle

\begin{abstract}

Advanced face swapping methods have achieved appealing results. However, most of these methods have many parameters and computations, which makes it challenging to apply them in real-time applications or deploy them on edge devices like mobile phones. In this work, we propose a lightweight Identity-aware Dynamic Network (IDN) for subject-agnostic face swapping by dynamically adjusting the model parameters according to the identity information. In particular, we design an efficient Identity Injection Module (IIM) by introducing two dynamic neural network techniques, including the weights prediction and weights modulation. Once the IDN is updated, it can be applied to swap faces given any target image or video. The presented IDN contains only 0.50M parameters and needs 0.33G FLOPs per frame, making it capable for real-time video face swapping on mobile phones. In addition, we introduce a knowledge distillation-based method for stable training, and a loss reweighting module is employed to obtain better synthesized results. Finally, our method achieves comparable results with the teacher models and other state-of-the-art methods.

\end{abstract}

\section{Introduction}
Recently, face swapping has drawn much attention from the research community, and it has many applications in visual effects. Face swapping means transferring the identity information of the source image to the target image while keeping the other attributes like the expression and background of the target image unchanged. Face swapping has achieved rapid progress with deep learning. However, most of these methods require many parameters and involve high computation costs. For example, FaceShifter \cite{li2019faceshifter} contains two stages for face swapping. The first stage alone has nearly 421M parameters and 97.4G FLOPs.
FSGAN \cite{nirkin2019fsgan} proposes a more complicated algorithm, which has over 226M parameters and 2440G FLOPs in total. Despite SimSwap \cite{chen2020simswap} claims that it presents an efficient network, it still has 107M parameters and 55.7G FLOPs.   
It is not only challenging to deploy these methods on edge devices, but these methods also require plenty of time for video face swapping, which means we cannot deploy them in real-time applications.

\begin{figure}[t!]
	\centering
	\includegraphics[width=0.48\textwidth]{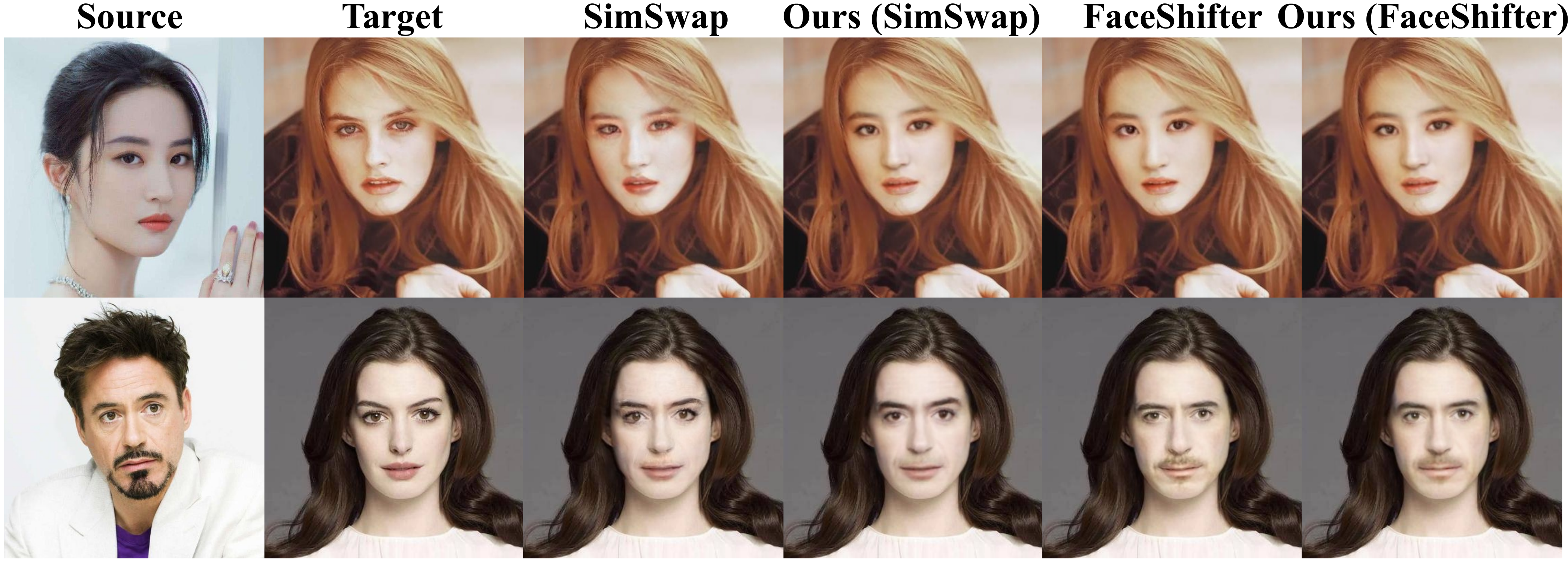}
	\caption{Face swapping results of our and the teacher models by swapping the face of the target image with the source image. Our method decreases the computations of SimSwap and FaceShifter by 146 and 207 times while preserving the visual fidelity.}
	\vspace{-1em}
	\label{fig:teaser}
\end{figure}

A natural idea to address the challenges above is using model compression techniques to produce a lightweight network for face swapping. However, when we apply the channel pruning technique and shrink the network to significantly compress the SimSwap or FaceShifter, we observe that it is troublesome to obtain pleasant face swapping results. In particular, some of the generated images have apparent artifacts, 
or the identities of the generated images may not be similar to the source images as the compressed model is insufficient to inject identity information with the limited network capacity. Inspired by the subject-aware face swapping technique \cite{perov2020deepfacelab}, we also find that the better swapped images can be obtained if we fix the source image and train a lightweight model for a specific identity. 
Therefore, to achieve subject-agnostic and real-time face swapping, an intuitive idea is to adjust the parameters of a neural network according to the identity information.

Inspired by the dynamic neural network techniques, we propose a lightweight Identity-aware Dynamic Network (IDN) for real time face swapping. To efficiently inject identity information, we also design an Identity Injection Module (IIM) using weights prediction \cite{de2016dynamic} and weights modulation \cite{karras2020analyzing} to adjust the parameters of IDN. In this way, the IDN can be updated given the needed identity information, and then we can achieve fast face swapping for any target image or video. 
Our method can significantly reduce the parameters and computations using these designs and achieve comparable results with other state-of-the-art methods. 
The proposed IDN has only 0.50M parameters and 0.33G FLOPs per frame for video face swapping when the input size is 224$\times$224. 
This means that we reduce the parameters and computations of the recent face swapping methods by more than 100 times. 
Without further optimization such as quantization, our model can achieve real-time face swapping on the mobile phone with MediaTek Dimensity 1100 chip, arriving at 26 FPS.

Generally, training a neural network for face swapping is unstable,
and the generated images may have obviously artifacts. 
We notice that if we transfer the face swapping to a paired training task using knowledge distillation \cite{hinton2015distilling}, 
we can achieve a more stable training process and get better results. 
Therefore, we employ a well-trained network as the teacher and train our lightweight network as the student. However, the teacher model may also produce some failure cases, such as the generated image having a low identity similarity with the source image. The student model can be misled by these failure cases and produce suboptimal results. In this paper, we propose a loss reweighting module to alleviate this problem. In particular, we evaluate the quality of the teacher outputs and adaptively adjust the weighting of distillation loss simultaneously. In this way, the student network can learn from better teacher outputs and therefore obtain better generated results with fewer artifacts and higher identity similarity.

The contributions of our paper are listed as follows:

1. We propose a real-time framework for video face swapping. It contains only 0.50M parameters and 0.33G FLOPs, and arrives at 26 FPS on the mobile phone.

2. We present an Identity Injection Module (IIM), which utilizes the weights prediction and weights modulation for more efficient identity information injection to build an Identity-aware Dynamic Network (IDN).

3. To stabilize the learning process, we train the proposed network using a knowledge distillation framework and propose a loss reweighting module to improve the generated results qualitatively and quantitatively.

\section{Related Work}
\subsubsection{Face swapping.}
Deep learning based face swapping has achieved significant improvement. The popular DeepFaceLab \cite{perov2020deepfacelab} trains an Encoder-Decoder for subject-aware face swapping and has achieved appealing results. However, this method only supports face swapping for two specific identities. Recently, additional subject-agnostic methods have been proposed that are more convenient to be employed. These methods can be roughly divided into two categories: source-oriented and target-oriented methods. 

The source-oriented methods first transfer the pose and expression of the source image to the target image, and then they apply a blending method to obtain the swapped face image. For example, \cite{nirkin2018face} employ 3DMM \cite{blanz1999morphable} to align the source image with the target image. FSGAN \cite{nirkin2019fsgan} trains two respective models to implement the face reenactment and blending process. However, these methods are unstable and prone to generating artifacts, such as unnatural color.
Target-oriented methods blend the features of the source image and target image to obtain the swapped face. FaceShifter \cite{li2019faceshifter} leverages a two-stage framework. The first stage is for face swapping, and the second stage is for occlusion processing. SimSwap \cite{chen2020simswap} utilizes AdaIN \cite{AdaIN} to inject the identity information into the target feature map. FaceController \cite{xu2021facecontroller} proposes a face representation based on 3DMM coefficients, as well as style and identity embedding, which can achieve more attribute editing than face swapping. HifiFace \cite{wang2021hififace} utilizes 3DMM coefficients to preserve the face shape of the source image for face swapping. 
However, the above methods have many parameters and computations that restrict the usage of these methods for face swapping.

\subsubsection{Dynamic neural networks.}
A dynamic neural network refers to one that adapts its structure or parameters to the input during inference, which can result in greater computational efficiency. This was first proposed by \cite{de2016dynamic}. This technique has been applied to other applications such as style transfer \cite{shen2018neural}, super-resolution \cite{jo2018deep, hu2019meta}, and image-to-image translation \cite{liu2019learning, SPADE}, etc.

\subsubsection{Knowledge distillation.}
Knowledge distillation \cite{hinton2015distilling} was proposed for transferring the knowledge in a larger teacher network to a smaller student network, which is widely used for model compression \cite{chen2017learning, wang2019private}. Recently, \cite{yim2017gift, li2020gan, jin2021teachers} use knowledge distillation to compress the generative adversarial network \cite{goodfellow2014generative} for image-to-image translation. 
These methods have also confirmed that utilizing knowledge distillation can improve the training stability for an unpaired training task by transferring it to a paired learning. Therefore, we introduce knowledge distillation into face swapping.

\section{Method}

\begin{figure*}[t!]
	\centering
	\includegraphics[width=0.99\textwidth]{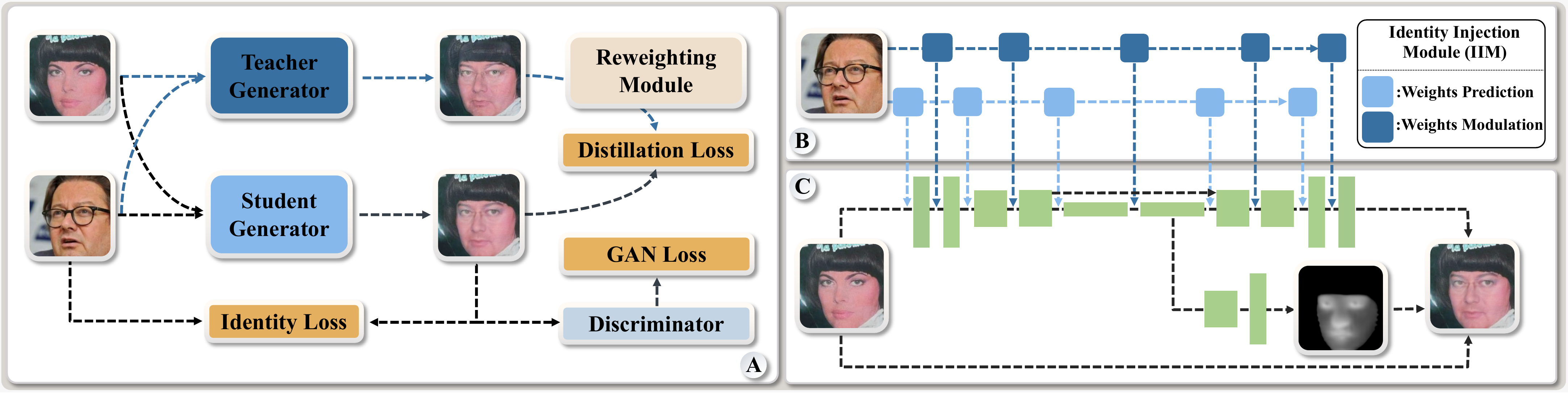}
	\caption{MobileFaceSwap framework: (a) The overall training process. (b) The Identity Injection Network (IIN) of MobileFaceSwap, which contains several Identity Injection Modules (IIM) and utilizes the identity information to predict or modulate the weights of the IDN. (c) The architecture of the Identity-aware Dynamic Network (IDN) and a weakly semantic fusion module for face swapping, that contain only 0.50M parameters and 0.33G FLOPs in total.}
	\label{fig:framework}
\end{figure*}

In this section, we will first describe the network architecture of our MobileFaceSwap, including the details of the Identity Injection Module (IIM), Identity-aware Dynamic Network (IDN), and the weakly semantic fusion module. Then, we introduce the knowledge distillation and our loss reweighting module to address the training stability problem and synthesis better swapped results. The overall framework is illustrated in Fig. \ref{fig:framework}.

\subsection{Network Architecture}

The main architecture of our method contains two neural networks. One is an Identity Injection Network (IIN) that has several Identity Injection Modules (IIM) to obtain the parameters required, while the other uses these parameters to construct a lightweight network named Identity-aware Dynamic Network (IDN) for the inference process. 
Given the identity representation of the source image by ArcFace \cite{deng2019arcface}, the IIM contains two dynamic neural network techniques to inject identity information to the IDN. Once we finish the injection process, we can swap faces with any target image or video using this lightweight IDN.

The IDN is simplified from U-Net \cite{ronneberger2015u} by replacing the standard convolution with the depthwise, and pointwise convolution \cite{chollet2017xception}. To modify the parameters of IDN according to a given source image, we introduce weights prediction \cite{de2016dynamic}, and weights modulation  \cite{karras2020analyzing} for depthwise and pointwise convolutions, respectively. Let $C_{in}$, $C_{out}$, and $K$ denote the input channels, output channels, and kernel size of a convolution layer, respectively. We utilize the identity embedding ${z_{id}}$ as the input to predict the weights of a depthwise convolution layer as follows:
\begin{equation}
\begin{aligned}
 \mathcal{W}_{d} = \mathcal{F}_{p}({z_{id}}),
\end{aligned}
\end{equation}
where $\mathcal{W}_{d}\in \mathbb{R}^{C_{in} \times 1 \times K \times K}$ represents the weights of a depthwise convolution layer in IDN, and $\mathcal{F}_{p}$ represents the prediction module that contains several convolution layers of the IIM.
Predicting the weights  $ \mathcal{W}_{p} \in \mathbb{R}^{C_{in} \times C_{out} \times 1 \times 1}$ for pointwise convolution needs more parameters since $C_{out}$ is generally far outweigh than $K \times K$.
Also, we observe that it can obtain better results by employing the weights modulation to inject the identity information into the weights of a pointwise convolution layer. Let $\mathcal{W}_{p}$, $\mathcal{\hat{W}}_{p}$, and $\mathcal{\tilde{W}}_{p}$ represent the origin, modulation, and demodulation weights of a pointwise convolution respectively, and $i$,  $j$, and $k$ enumerate the input/output feature maps and spatial footprint of the convolution, respectively. Then, the weights modulation technique is formulated as follows:
\begin{equation}
\begin{aligned}
\mathcal{\hat{W}}_{p}^{(i, j, k)} &= \mathcal{F}_m(z_{id})^{(i)} \cdot \mathcal{W}_{p}^{(i, j, k)}, \\
\mathcal{\tilde{W}}_{p}^{(i, j, k)} &= \frac{\mathcal{\hat{W}}_{p}^{(i, j, k)}}{\sqrt{\sum_{i,k}(\mathcal{\hat{W}}_{p}^{(i, j, k)}) ^ 2 + \epsilon}},
\end{aligned}
\end{equation}
where $\mathcal{F}_{m}$ is the modulation module of the IIM, which includes several fully-connected layers,
and $\epsilon$ is used for number stability. By applying IIM, the IDN can achieve subject-agnostic face swapping, which contains 0.413M parameters and 0.328G FLOPs only.

In practice, it is hard to keep the background and hair of the generated images being the same as the target images.
This problem also causes jitter for video face swapping. Generally, it can be solved by adding computationally intensive post-processing such as face segmentation \cite{yu2018bisenet}. In this paper, we propose a weakly semantic fusion module to merge the background of the target image. We predict a weakly fusion mask by reusing the feature maps of the IDN. Our semantic fusion module contains only 0.083M parameters and 0.005G FLOPs. Even though our semantic fusion module is exceptionally lightweight, the predicted fusion mask is pretty well as demonstrate in Fig. \ref{fig:framework}. Finally, we have built the entire network for video face swapping, which has 0.495M parameters and needs 0.333G FLOPs in total when the input size is 224$\times$224.

\subsection{Training Objectives}
Generally, training a face swapping network requires many loss functions to guarantee that the generated result meets the definition of face swapping. The competition of these different losses makes the training process unstable and easier to generate artifacts as there is no paired ground truth for the constraint. In this paper, we transfer face swapping to a paired training by introducing a knowledge distillation  \cite{hinton2015distilling} framework. Given a well-trained face swapping network as the teacher and the proposed network as the student, we utilize L1 loss and perceptual loss \cite{johnson2016perceptual} between the student output $I_g$ and the teacher output $I_g'$ as follows:
\begin{equation}
\begin{aligned}
\label{dist_loss}
\mathcal{L}_{rec} &= || {I_g'} - I_{g} ||, \\
\mathcal{L}_{per} &= \sum_{i=1}^{L} \frac{1}{N_i}  || \mathcal{F}_{VGG}^{(i)}(I_g') - \mathcal{F}_{VGG}^{(i)}(I_g) ||^2,
\end{aligned}
\end{equation}
where $\mathcal{F}_{VGG}^{(i)}$ denotes the $i$-th layer
with $N_i$ elements of the VGG network \cite{simonyan2014very}.
Our method can achieve more stable training process and better synthesized results by knowledge distillation. 

However, the teacher is not perfect and some bad cases can be found in the teacher outputs. Specifically, we roughly divide these failure cases into two categories of face swapping. First, 
some teacher outputs cannot keep the identity well of the source images. Second, some teacher outputs may have unnatural results or artifacts, such as noise and a dirty forehead. If we assign an equal weight for each teacher output in Equ. \ref{dist_loss}, the student network will also learn to generate these bad cases. In this paper, we propose a loss reweighting module to alleviate this problem. We consider the identity similarity and image quality for each teacher output. Specifically, we use the square of the cosine distance between the identity representations of the teacher output and source image to measure the identity similarity. 
However, evaluating the quality of the image is non-trivial. Since we observe that the swapped results gradually become better during the training process, to assess the image quality of the teacher outputs properly, we assign each teacher output with a score between 0 to 1, according to the percentage of the completed iterations in the teacher training process. Subsequently, we use a ResNet-18 \cite{he2016deep} to regress these scores by supervision with L2 Loss. Finally, we employ this model $Q$ to evaluate the image quality of the teacher outputs. The final distillation loss reweighting coefficient $\alpha$ is calculated as follows:
\begin{equation}
\begin{aligned}
    \alpha = Cos(z_{id}, {z_{id}'}) ^ 2 \times Q({I_g'}),
\end{aligned}
\end{equation}
where $z_{id}'$ denotes the identity representation of the teacher output.
By introducing the loss reweighting module, not only can we improve the identity similarity between the generated image and the source image, but we can also improve the quality of swapped results.

\begin{figure}[t!]
	\centering
	\includegraphics[width=0.47\textwidth]{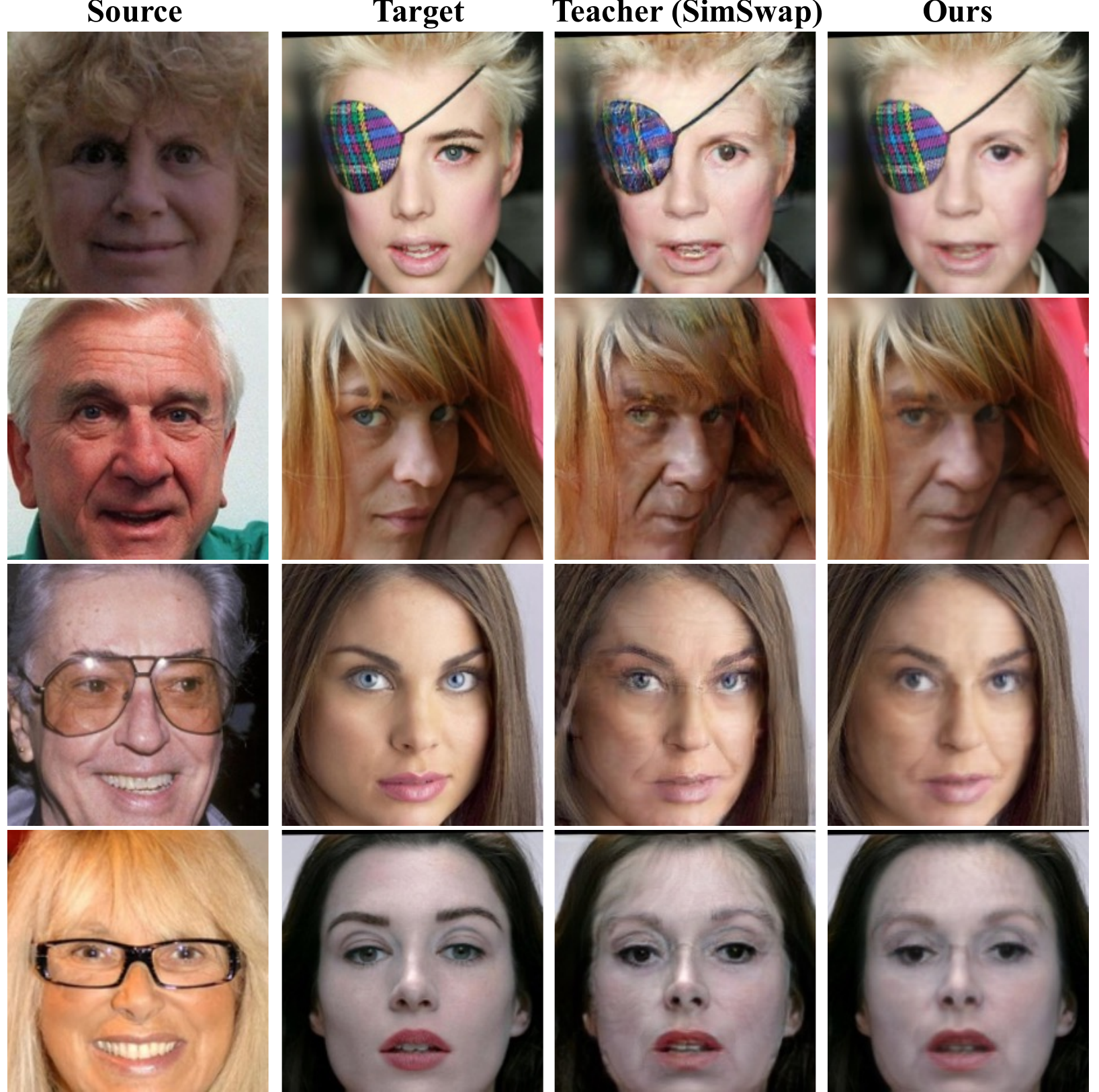}
	\caption{Comparison with the teacher (SimSwap) results.}
	\label{fig:compare_with_simswap}
\end{figure}

Although we transfer the teacher knowledge to the student model using distillation loss, the supervision on identity is insufficient. The quantitative results about identity similarity drop significantly compared with the teacher. Therefore, we add a supplementary identity loss for better identity supervision following FaceShifter \cite{li2019faceshifter}. We employ ArcFace \cite{deng2019arcface} as the extractor $\mathcal{F}_{id}$ and calculate the cosine similarity between the identity representations of the source image $I_s$ and generated image $I_g$.
\begin{equation}
\begin{aligned}
    \mathcal{L}_{id} = 1 - Cos(\mathcal{F}_{id}(I_s), 
    \mathcal{F}_{id}(I_g)),
\end{aligned}
\end{equation}
For the fusion mask prediction, we employ weak supervision. Specifically, we only constrain the background of the generated image to keep it being the same as the target image. Compared with full supervision, our weak mask loss can better retain the attributes of the target image, such as local textures. The mask loss is defined as follows:
\begin{equation}
\begin{aligned}
    \mathcal{L}_{mask} = ||M_t[bg] - M_g[bg]||,
\end{aligned}
\end{equation}
where $M_t$ and $M_g$ represent the mask of the target and generated image, respectively, $M[bg]$ denotes the background elements of the mask. We use \cite{yu2018bisenet} to obtain $M_t$ by combing the semantic labels corresponding to the background.

The total loss is defined as a sum of the above losses.
\begin{equation}
\begin{aligned}
    \mathcal{L} = \mathcal{L}_{adv}  +  \alpha ( \lambda_{rec} \mathcal{L}_{rec} + \lambda_{per} \mathcal{L}_{per})  \\
    +  \lambda_{id} \mathcal{L}_{id} + \lambda_{mask} \mathcal{L}_{mask},
\end{aligned}
\end{equation}
where $\mathcal{L}_{adv}$ denotes the GAN loss, and we set $\lambda_{id}=3$, $\lambda_{rec}=30$, $\lambda_{per}=5$, and $\lambda_{mask}=10$, respectively.
\begin{figure}[t!]
	\centering
	\includegraphics[width=0.47\textwidth]{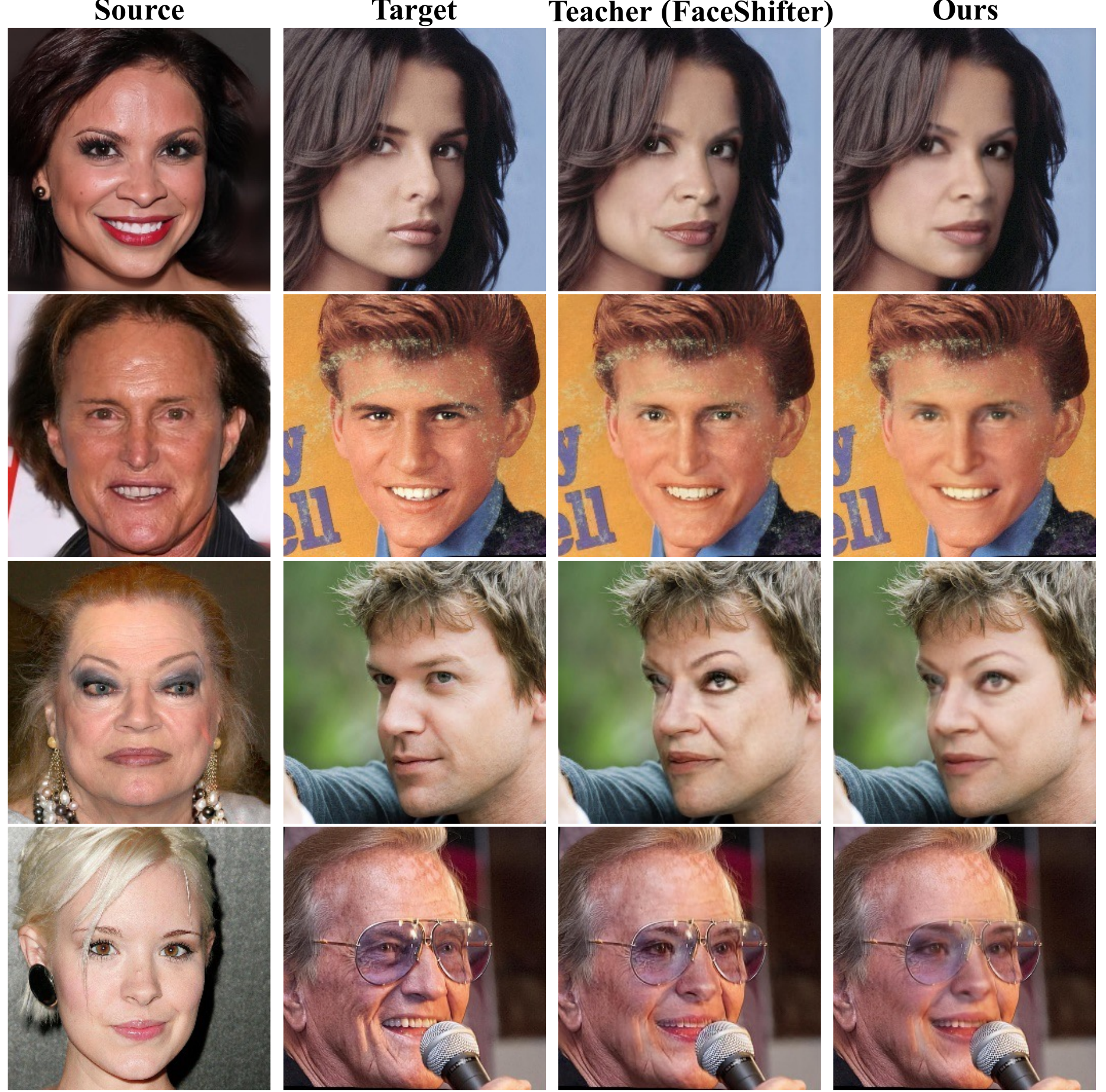}
	\caption{Comparison with the teacher (FaceShifter) results.}
	\label{fig:compare_with_faceshifter}
\end{figure}

\section{Experiments}

\begin{figure*}[t!]
	\centering
	\includegraphics[width=0.99\textwidth]{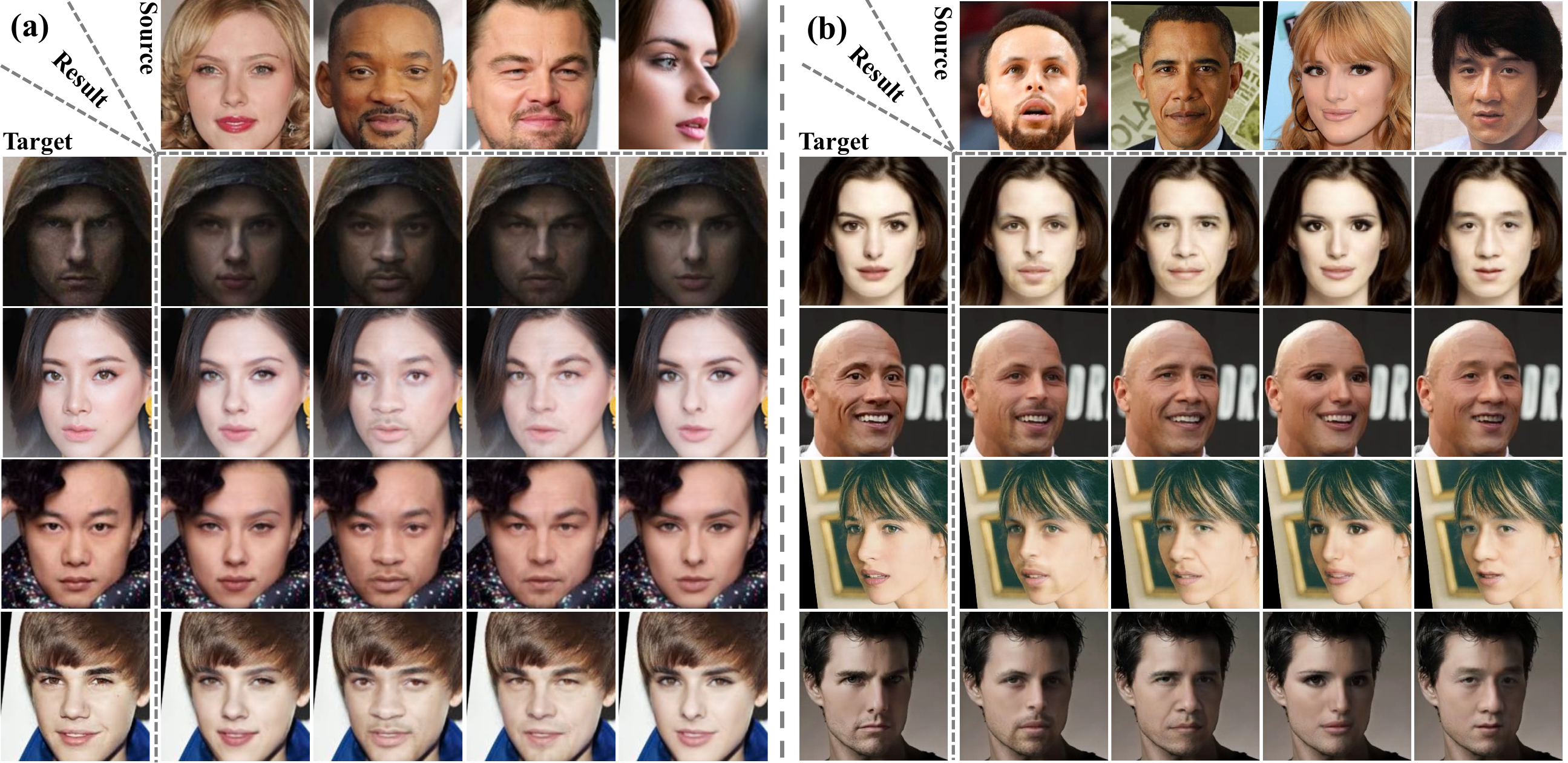}
	\caption{Further face swapping results of our models that are distilled from SimSwap (a) and FaceShifter (b) respectively. }
	\label{fig:demo_many}
\end{figure*}
\subsubsection{Implementation details.}
The training images are collected from VGGFace2 \cite{cao2018vggface2}. We select the landmarks between the two eyes larger than 70 pixels and get 550K images. We conduct experiments on two famous face swapping algorithms as the teacher models, including SimSwap \cite{chen2020simswap} and FaceShifter \cite{li2019faceshifter}. The student model uses the same face alignment algorithm as the teacher model.
The image sizes are 224$\times$224 and 256$\times$256 for SimSwap and FaceShifter, respectively.


\subsection{Qualitative Results}
\subsubsection{Comparison with teachers.}
First, we compare our method with the teacher methods. The model of SimSwap is derived from the official repository, while the FaceShifter has been implemented by ourselves since the source codes are not released. The test images are collected from CelebA-HQ \cite{karras2018progressive}. As shown in Fig. \ref{fig:compare_with_simswap}, our results have fewer artifacts than those of SimSwap. Specifically, employing our semantic fusion module, our method can keep the hair and background more stable than SimSwap, as shown in the first and second rows of Fig. \ref{fig:compare_with_simswap}. In addition, The result of SimSwap has artifacts around the eyes when the subject of the source image wears glasses, while our method can obtain clearer results. We demonstrate the comparison with FaceShifter in Fig. \ref{fig:compare_with_faceshifter}. As shown in the first row, our method can keep the expression of the target image better than FaceShifter. At the same time, our method can also generate impressive results under uncommon conditions like occlusions or large poses.
\subsubsection{Demonstration.}
We demonstrate further results of our models in Fig. \ref{fig:demo_many}. These superstar images are collected from the internet. As we can see, our models generate appealing results for different types of source and target images. 
\subsubsection{Comparison with state-of-the-art methods.}
\begin{figure*}[t!]
	\centering
	\includegraphics[width=0.99\textwidth]{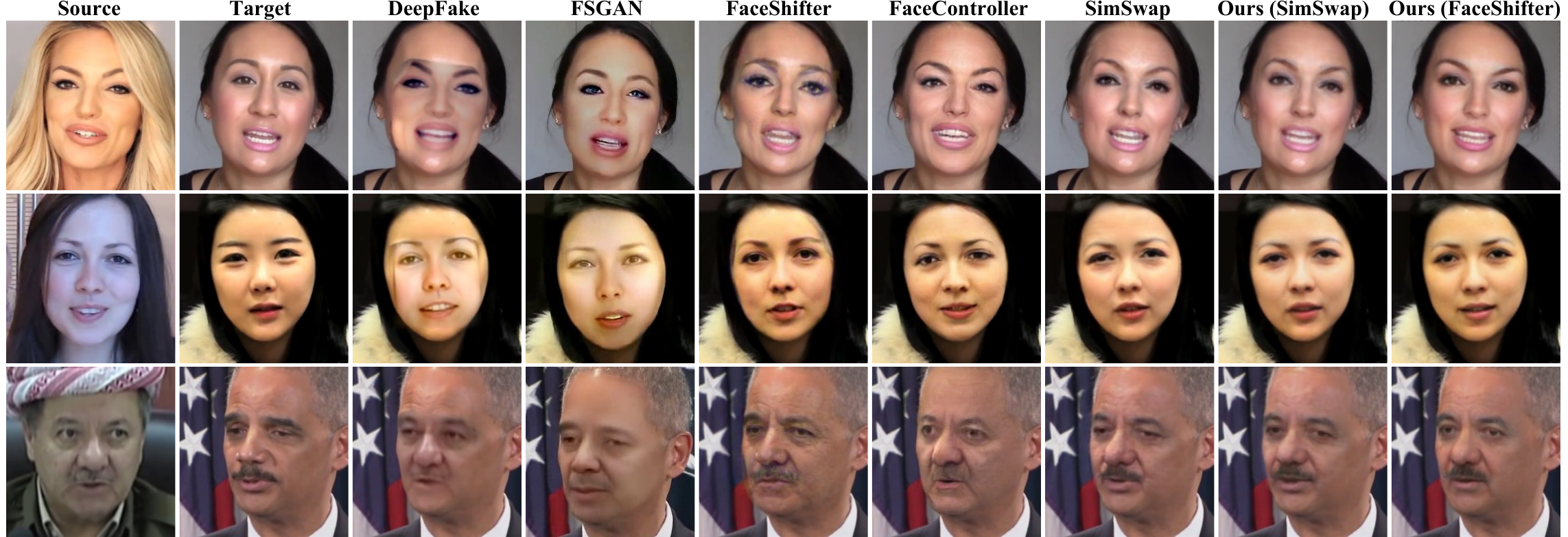}
	\caption{Comparison with DeepFakes, FSGAN, FaceShifter, FaceController, and SimSwap on the FaceForensics++ dataset.}
	\label{fig:faceForensic_comparsion}
\end{figure*}

\begin{figure}[t!]
	\centering
	\includegraphics[width=0.49\textwidth]{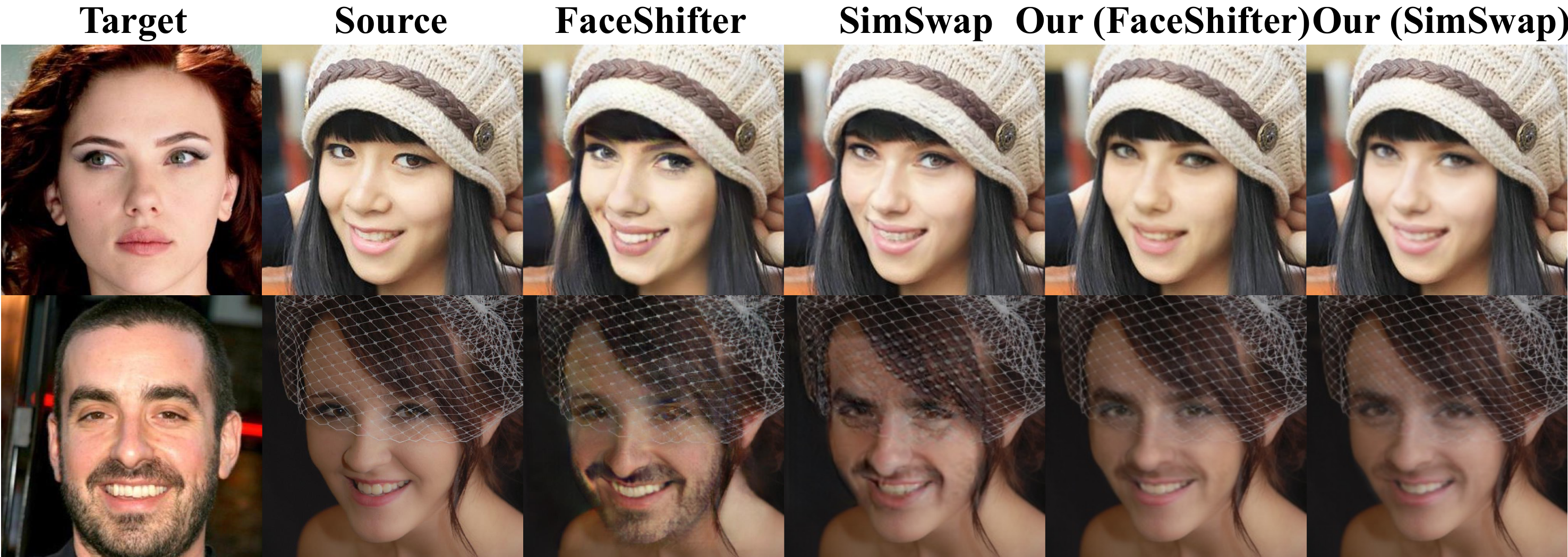}
	\caption{More comparison with SimSwap and FaceShifter (Zoom view better).}
	\label{fig:more_faceshifter_compare}
\end{figure}

We also compare our method with other face swapping methods, including DeepFakes\footnote{\url{https://github.com/deepfakes/faceswap}\label{deepfake}}, FSGAN \cite{nirkin2019fsgan}, FaceShifter \cite{li2019faceshifter}, SimSwap \cite{chen2020simswap}, and recently proposed FaceController \cite{xu2021facecontroller} on the FaceForensics++ \cite{rossler2019faceforensics++} dataset, which is a widely used dataset for deepfakes creation and detection. For a fair comparison, we use the official results that are derived from the FaceForensics++ dataset for DeepFakes and FaceShifter. While for FSGAN and SimSwap, these results are generated by the official codes and models. The results of FaceController are cropped from its paper. The comparison results are shown in Fig. \ref{fig:faceForensic_comparsion}. 
As we can see, the results of DeepFakes and FSGAN have noticeable artifacts like unnatural color.
Compared with FaceController, our method can keep the expression of the target image better. For SimSwap and FaceShifter, our method achieves comparable results at these images. We also compare our method with SimSwap and FaceShifter using the images from FaceShifter for more qualitative results as shown in Fig. \ref{fig:more_faceshifter_compare}.  Our method can retain the attributes of the target image better than FaceShifter, and generate fewer artifacts than SimSwap.
\begin{table*}[t!]
\centering
\small
\begin{tabular}{lcccccccc}
\hline
Method       &  Size &  (I) Params (M)  &  (I) FLOPs (G)  &  (V) Params (M)   &  (V) FLOPs (G) & (V) FPS  & Id$\uparrow$ & Pose$\downarrow$ \\ \hline
DeepFakes    &  64   & 82.1   &  \textbf{1.90}   &   82.1     &    1.90     & 9.5 &      81.96     &   4.14   \\
FSGAN        & 256   &  226     &    2440     &      226     &     2240   & -  &    57.34      &  3.81 \\ 
SimSwap      &  224  &  107 &  55.7   &    45.6    &     48.2   &  0.64   &        92.83    &  1.53         \\
FaceShifter  &  256  &  421   &  97.4   &    350     &    91.1     &  - &        97.38      &  2.96\\
FaceController& 224  &  306   &  192    &    236     &      177  &  - &        \textbf{98.27}      &  2.65  \\ \hline
Teacher (SimSwap) & 224 & 107  &  55.7    &     45.6   &   48.2   &  0.64  &    95.94             &   1.39           \\
Teacher (FaceShifter) & 256 & 421 &  97.4   &       350  &    91.1 & -   &     97.15         &   1.76  \\ \hline
Ours (SimSwap)& 224 &    \textbf{72.8} &  8.07 & \textbf{0.50} &      \textbf{0.33} &   \textbf{25.6} &     95.98     &  \textbf{1.32}   \\
Ours (FaceShifter)& 256 &    \textbf{72.8} & 8.18  &  \textbf{0.50} &      0.44 & 19.7 &    96.10        &   1.70  \\ \hline
Id Network  & 112 & 52.2  &  7.52    &     -   &   -      &    -             &   -       & -    \\ \hline
\end{tabular}
\caption{The comparison of different face swapping methods, where I and V represent the image and video scenes, respectively. 
Note that we count the parameters and computations of the Id Network for image face swapping if these algorithms need it, but not for video scenes.
The Size means the size of the input image. 
FPS is tested under the mobile phone with MediaTek Dimensity 1100 chip. In the last two columns, we report the accuracies of identity retrieval and pose errors on the FaceForensics++ dataset.}
\label{tab:quantitative_comparison}
\end{table*}

\subsection{Quantitative Results}
We compare our method with other methods about parameters and computations in more detail as shown in Table \ref{tab:quantitative_comparison}. We report the parameters and computations at image-based and video-based conditions, respectively. As we can see, our method has significant advantages in relation to the parameters and computations. Specifically, for video face swapping, our method contains only 0.50M parameters and 0.44G FLOPs when the input size is 256$\times$256, which are fewer than 100 times as compared to other face swapping methods like SimSwap and FaceShifter. For image-based face swapping, most of the parameters and computations of our method are consumed by the identity network \cite{deng2019arcface}. DeepFakes also has advantages with computations. However, it requires training a different model when given a new identity, and the size of the input is 64$\times$64. Nevertheless, our method still has fewer computations than DeepFakes for video face swapping. Then, we test our approach and SimSwap at the mobile phone with the MediaTek Dimensity 1100 chip to demonstrate our superiority in speed. Without further optimizations, our method arrives at 25.6 FPS, 40 times faster than SimSwap. We did not test others methods as these models are too heavy for mobile phones.

The quantitative comparisons with respect to the quality of face swapping are shown in the last two columns of Table \ref{tab:quantitative_comparison}.  We follow the evaluation metrics used in FaceShifter and SimSwap, including identity retrieval accuracy and posture similarity on the FaceForensics++ dataset. 
For identity retrieval, we use the CosFace \cite{wang2018cosface} to extract the identity embedding and retrieve the closest face using cosine similarity. 
For pose evaluation, we employ a pose estimator \cite{ruiz2018fine} to estimate head pose and then report the L2 distance of pose vectors. Note that the results of Teacher (SimSwap) are different from that reported in the original SimSwap paper since the authors released a different model in their official repository. As we can see, the obtained models using our method can achieve comparable results with their teachers. The identity retrieval accuracy and pose similarity of our results are better than the Teacher (SimSwap). Compared with other methods, our method can better balance identity retrieval accuracy and pose similarity.

\subsection{Ablation Study}

\begin{figure*}[t!]
	\centering
	\includegraphics[width=0.98\textwidth]{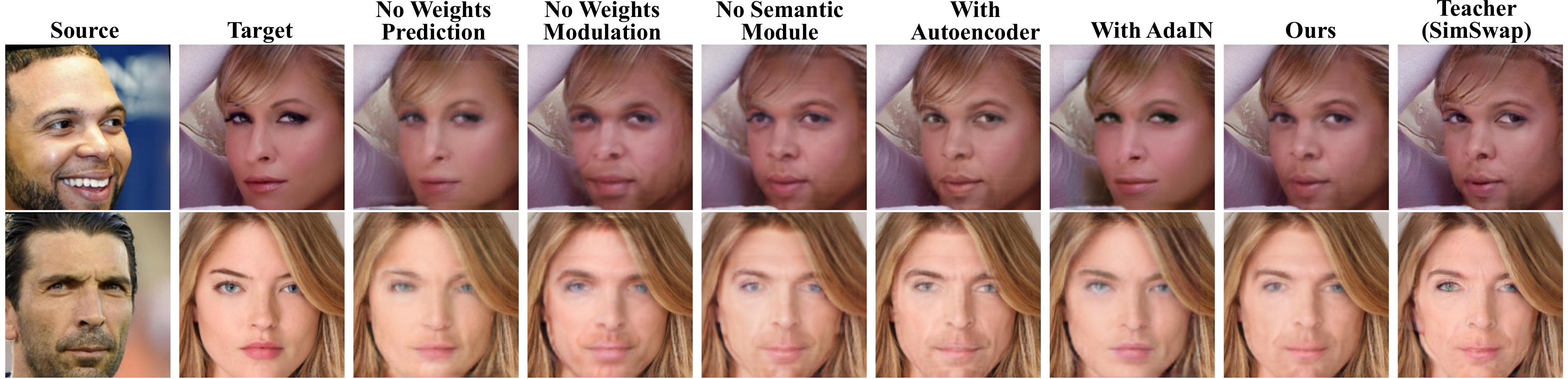}
	\caption{Ablation study about network architecture.}
	\label{fig:ablation_architecture}
\end{figure*}
\begin{figure*}[t!]
	\centering
	\includegraphics[width=0.98\textwidth]{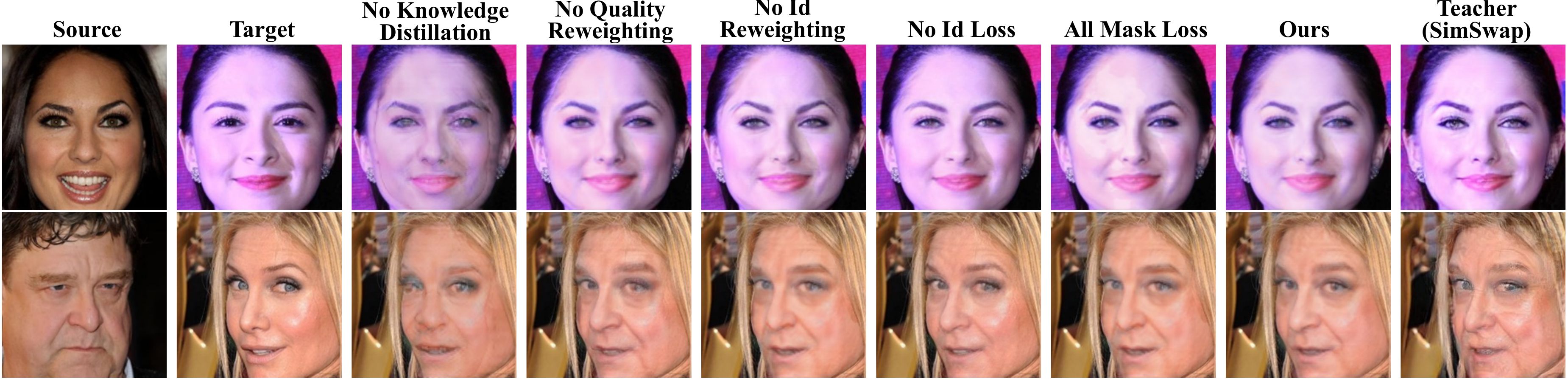}
	\caption{Ablation study about training objectives.}
	\label{fig:ablation_loss}
\end{figure*}

\subsubsection{Effectiveness of the network.}
We conduct all ablation studies using SimSwap as the teacher model to verify the efficiency and necessity of our network architecture designs. 
The qualitative and quantitative results are shown in Fig. \ref{fig:ablation_architecture} and Table \ref{tab:ablation_quan}, respectively. First, suppose we drop out the weights prediction or replace the weights modulation with AdaIN \cite{AdaIN}, the revised method can be regarded as a compression version of SimSwap. In these cases, the training processes are unstable and cannot generate acceptable swapping results. The identity retrieval accuracy significantly drops as shown in Table \ref{tab:ablation_quan}. If we drop out of the semantic module, there is a clear difference in the backgrounds between the generated and target image, which is similar to SimSwap. Then, the qualitative and quantitative results drop slightly by removing the weighting modulation or replacing the U-Net-based with AutoEncoder-based network, which results in more artifacts than ours.
\begin{table}[t!]
\centering
\begin{tabular}{lcccc}
\hline
Method                  &    Id$\uparrow$    & Pose$\downarrow$   & FID$\downarrow$ \\ \hline
No Weights Prediction    &       16.29           &    1.40          &  16.75        \\
No Weights Modulation    &      86.20       &    2.12          &  9.92        \\ 
No Semantic Module      &     \textbf{96.10}      &    1.69          &  12.13       \\ 
With AutoEncoder        &       92.78           &    1.43          &  7.77         \\
With AdaIN              &        18.34          &    1.30         &  12.04        \\ \hline
No Knowledge Distillation     &        88.48      &       3.21    &      17.37     \\
No Quality Reweight           &        95.28      &       1.50    &      8.12      \\ 
No Id Reweight                &        93.71      &       1.52    &      7.38      \\
No Id Loss                    &        28.13      & \textbf{0.98}  & \textbf{4.76}    \\
All Mask Loss                 &        95.46      &       1.42    &      7.44      \\ \hline
Teacher (SimSwap)             &        95.94      &    1.39       &  10.26        \\ \hline
Ours                          &        95.98      &      1.32     &  6.79        \\ \hline
\end{tabular}
\caption{Ablation study results of different network architectures and training objectives on the FaceForensics++ dataset.}
\label{tab:ablation_quan}
\end{table}
\subsubsection{Effectiveness of training objectives.}

To verify the advantages of the training objectives, we have conducted many ablation studies on loss functions, and the results have been shown in Fig. \ref{fig:ablation_loss} and 
Table \ref{tab:ablation_quan}.
First, if we drop out the knowledge distillation, the generated results have noticeable artifacts and worse quantitative results. Then, we evaluate our loss reweighting module, including the identity reweighting and image quality reweighting. 
The identity retrieval accuracy increases by 2.27\% when adding the identity reweighting module. When adding image quality reweighting, our model can generate slightly better results. Then, if we drop out of the identity loss, the generated results have fewer artifacts than those of the proposed method. However, the identity retrieval accuracy is also pretty low. Last, as demonstrated in Fig. \ref{fig:ablation_loss}, we can see that using a weakly semantic module can clearly decrease the color difference between the generated image and the target image.

\section{Conclusion}

In this work, we propose MobileFaceSwap for real-time subject-agnostic face swapping. We design an efficient Identity Injection Module (IIM) to adjust the parameters of the Identity-aware Dynamic Network (IDN) adaptively. Then, we use the knowledge distillation and design a loss reweighting module to obtain better swapped results. Our method can be deployed on mobile phones, perform real-time face swapping. 
Besides, we can generate some forgery samples by MobileFaceSwap and hope these will have a little impact on forgery detection, as some forgery techniques are likely to be abused for malicious purposes.

\section{Acknowledgement}
This work is supported by CCF-Baidu Open Fund.

{\fontsize{9.0pt}{10.0pt} \selectfont
\bibliography{main}

\begin{thebibliography}{35}
\providecommand{\natexlab}[1]{#1}

\bibitem[{Blanz and Vetter(1999)}]{blanz1999morphable}
Blanz, V.; and Vetter, T. 1999.
\newblock A morphable model for the synthesis of 3D faces.
\newblock In \emph{Proceedings of the 26th annual conference on Computer
  graphics and interactive techniques}, 187--194.

\bibitem[{Cao et~al.(2018)Cao, Shen, Xie, Parkhi, and
  Zisserman}]{cao2018vggface2}
Cao, Q.; Shen, L.; Xie, W.; Parkhi, O.~M.; and Zisserman, A. 2018.
\newblock Vggface2: A dataset for recognising faces across pose and age.
\newblock In \emph{13th IEEE international conference on automatic face \&
  gesture recognition}, 67--74. IEEE.

\bibitem[{Chen et~al.(2017)Chen, Choi, Yu, Han, and
  Chandraker}]{chen2017learning}
Chen, G.; Choi, W.; Yu, X.; Han, T.; and Chandraker, M. 2017.
\newblock Learning efficient object detection models with knowledge
  distillation.
\newblock In \emph{Proceedings of the 31st International Conference on Neural
  Information Processing Systems}, 742--751.

\bibitem[{Chen et~al.(2020)Chen, Chen, Ni, and Ge}]{chen2020simswap}
Chen, R.; Chen, X.; Ni, B.; and Ge, Y. 2020.
\newblock SimSwap: An Efficient Framework For High Fidelity Face Swapping.
\newblock In \emph{Proceedings of the 28th ACM International Conference on
  Multimedia}, 2003--2011.

\bibitem[{Chollet(2017)}]{chollet2017xception}
Chollet, F. 2017.
\newblock Xception: Deep learning with depthwise separable convolutions.
\newblock In \emph{Proceedings of the IEEE Conference on Computer Vision and
  Pattern Recognition}, 1251--1258.

\bibitem[{De~Brabandere et~al.(2016)De~Brabandere, Jia, Tuytelaars, and
  Van~Gool}]{de2016dynamic}
De~Brabandere, B.; Jia, X.; Tuytelaars, T.; and Van~Gool, L. 2016.
\newblock Dynamic filter networks.
\newblock In \emph{Proceedings of the 30th International Conference on Neural
  Information Processing Systems}, 667--675.

\bibitem[{Deng et~al.(2019)Deng, Guo, Xue, and Zafeiriou}]{deng2019arcface}
Deng, J.; Guo, J.; Xue, N.; and Zafeiriou, S. 2019.
\newblock Arcface: Additive angular margin loss for deep face recognition.
\newblock In \emph{Proceedings of the IEEE Conference on Computer Vision and
  Pattern Recognition}, 4690--4699.

\bibitem[{Goodfellow et~al.(2014)Goodfellow, Pouget-Abadie, Mirza, Xu,
  Warde-Farley, Ozair, Courville, and Bengio}]{goodfellow2014generative}
Goodfellow, I.~J.; Pouget-Abadie, J.; Mirza, M.; Xu, B.; Warde-Farley, D.;
  Ozair, S.; Courville, A.; and Bengio, Y. 2014.
\newblock Generative adversarial nets.
\newblock In \emph{Proceedings of the 27th International Conference on Neural
  Information Processing}, 2672--2680.

\bibitem[{He et~al.(2016)He, Zhang, Ren, and Sun}]{he2016deep}
He, K.; Zhang, X.; Ren, S.; and Sun, J. 2016.
\newblock Deep residual learning for image recognition.
\newblock In \emph{Proceedings of the IEEE Conference on Computer Vision and
  Pattern Recognition}, 770--778.

\bibitem[{Hinton, Vinyals, and Dean(2015)}]{hinton2015distilling}
Hinton, G.; Vinyals, O.; and Dean, J. 2015.
\newblock Distilling the knowledge in a neural network.
\newblock \emph{arXiv preprint arXiv:1503.02531}.

\bibitem[{Hu et~al.(2019)Hu, Mu, Zhang, Wang, Tan, and Sun}]{hu2019meta}
Hu, X.; Mu, H.; Zhang, X.; Wang, Z.; Tan, T.; and Sun, J. 2019.
\newblock Meta-SR: A magnification-arbitrary network for super-resolution.
\newblock In \emph{Proceedings of the IEEE Conference on Computer Vision and
  Pattern Recognition}, 1575--1584.

\bibitem[{Huang and Belongie(2017)}]{AdaIN}
Huang, X.; and Belongie, S.~J. 2017.
\newblock Arbitrary Style Transfer in Real-Time with Adaptive Instance
  Normalization.
\newblock In \emph{Proceedings of the IEEE International Conference on Computer
  Vision}, 1510--1519.

\bibitem[{Jin et~al.(2021)Jin, Ren, Woodford, Wang, Yuan, Wang, and
  Tulyakov}]{jin2021teachers}
Jin, Q.; Ren, J.; Woodford, O.~J.; Wang, J.; Yuan, G.; Wang, Y.; and Tulyakov,
  S. 2021.
\newblock Teachers Do More Than Teach: Compressing Image-to-Image Models.
\newblock In \emph{Proceedings of the IEEE Conference on Computer Vision and
  Pattern Recognition}, 13600--13611.

\bibitem[{Jo et~al.(2018)Jo, Oh, Kang, and Kim}]{jo2018deep}
Jo, Y.; Oh, S.~W.; Kang, J.; and Kim, S.~J. 2018.
\newblock Deep video super-resolution network using dynamic upsampling filters
  without explicit motion compensation.
\newblock In \emph{Proceedings of the IEEE Conference on Computer Vision and
  Pattern Recognition}, 3224--3232.

\bibitem[{Johnson, Alahi, and Fei-Fei(2016)}]{johnson2016perceptual}
Johnson, J.; Alahi, A.; and Fei-Fei, L. 2016.
\newblock Perceptual losses for real-time style transfer and super-resolution.
\newblock In \emph{European conference on computer vision}, 694--711. Springer.

\bibitem[{Karras et~al.(2018)Karras, Aila, Laine, and
  Lehtinen}]{karras2018progressive}
Karras, T.; Aila, T.; Laine, S.; and Lehtinen, J. 2018.
\newblock Progressive Growing of GANs for Improved Quality, Stability, and
  Variation.
\newblock In \emph{International Conference on Learning Representations}.

\bibitem[{Karras et~al.(2020)Karras, Laine, Aittala, Hellsten, Lehtinen, and
  Aila}]{karras2020analyzing}
Karras, T.; Laine, S.; Aittala, M.; Hellsten, J.; Lehtinen, J.; and Aila, T.
  2020.
\newblock Analyzing and improving the image quality of stylegan.
\newblock In \emph{Proceedings of the IEEE Conference on Computer Vision and
  Pattern Recognition}, 8110--8119.

\bibitem[{Li et~al.(2020{\natexlab{a}})Li, Bao, Yang, Chen, and
  Wen}]{li2019faceshifter}
Li, L.; Bao, J.; Yang, H.; Chen, D.; and Wen, F. 2020{\natexlab{a}}.
\newblock Advancing high fidelity identity swapping for forgery detection.
\newblock In \emph{Proceedings of the IEEE Conference on Computer Vision and
  Pattern Recognition}, 5074--5083.

\bibitem[{Li et~al.(2020{\natexlab{b}})Li, Lin, Ding, Liu, Zhu, and
  Han}]{li2020gan}
Li, M.; Lin, J.; Ding, Y.; Liu, Z.; Zhu, J.-Y.; and Han, S. 2020{\natexlab{b}}.
\newblock Gan compression: Efficient architectures for interactive conditional
  gans.
\newblock In \emph{Proceedings of the IEEE Conference on Computer Vision and
  Pattern Recognition}, 5284--5294.

\bibitem[{Liu et~al.(2019)Liu, Yin, Shao, Wang, and Li}]{liu2019learning}
Liu, X.; Yin, G.; Shao, J.; Wang, X.; and Li, H. 2019.
\newblock Learning to predict layout-to-image conditional convolutions for
  semantic image synthesis.
\newblock In \emph{Proceedings of the 33rd International Conference on Neural
  Information Processing Systems}, 570--580.

\bibitem[{Nirkin, Keller, and Hassner(2019)}]{nirkin2019fsgan}
Nirkin, Y.; Keller, Y.; and Hassner, T. 2019.
\newblock F{SGAN}: Subject Agnostic Face Swapping and Reenactment.
\newblock In \emph{Proceedings of the IEEE International Conference on Computer
  Vision}, 7184--7193.

\bibitem[{Nirkin et~al.(2018)Nirkin, Masi, Tuan, Hassner, and
  Medioni}]{nirkin2018face}
Nirkin, Y.; Masi, I.; Tuan, A.~T.; Hassner, T.; and Medioni, G. 2018.
\newblock On face segmentation, face swapping, and face perception.
\newblock In \emph{2018 13th IEEE International Conference on Automatic Face \&
  Gesture Recognition (FG 2018)}, 98--105. IEEE.

\bibitem[{Park et~al.(2019)Park, Liu, Wang, and Zhu}]{SPADE}
Park, T.; Liu, M.; Wang, T.; and Zhu, J. 2019.
\newblock Semantic Image Synthesis With Spatially-Adaptive Normalization.
\newblock In \emph{Proceedings of the IEEE Conference on Computer Vision and
  Pattern Recognition}, 2337--2346.

\bibitem[{Perov et~al.(2020)Perov, Gao, Chervoniy, Liu, Marangonda, Um{\'e},
  Dpfks, Facenheim, RP, Jiang et~al.}]{perov2020deepfacelab}
Perov, I.; Gao, D.; Chervoniy, N.; Liu, K.; Marangonda, S.; Um{\'e}, C.; Dpfks,
  M.; Facenheim, C.~S.; RP, L.; Jiang, J.; et~al. 2020.
\newblock Deepfacelab: A simple, flexible and extensible face swapping
  framework.
\newblock \emph{arXiv preprint arXiv:2005.05535}.

\bibitem[{Ronneberger, Fischer, and Brox(2015)}]{ronneberger2015u}
Ronneberger, O.; Fischer, P.; and Brox, T. 2015.
\newblock U-net: Convolutional networks for biomedical image segmentation.
\newblock In \emph{International Conference on Medical image computing and
  computer-assisted intervention}, 234--241. Springer.

\bibitem[{Rossler et~al.(2019)Rossler, Cozzolino, Verdoliva, Riess, Thies, and
  Nie{\ss}ner}]{rossler2019faceforensics++}
Rossler, A.; Cozzolino, D.; Verdoliva, L.; Riess, C.; Thies, J.; and
  Nie{\ss}ner, M. 2019.
\newblock Faceforensics++: Learning to detect manipulated facial images.
\newblock In \emph{Proceedings of the IEEE International Conference on Computer
  Vision}, 1--11.

\bibitem[{Ruiz, Chong, and Rehg(2018)}]{ruiz2018fine}
Ruiz, N.; Chong, E.; and Rehg, J.~M. 2018.
\newblock Fine-grained head pose estimation without keypoints.
\newblock In \emph{Proceedings of the IEEE conference on computer vision and
  pattern recognition workshops}, 2074--2083.

\bibitem[{Shen, Yan, and Zeng(2018)}]{shen2018neural}
Shen, F.; Yan, S.; and Zeng, G. 2018.
\newblock Neural style transfer via meta networks.
\newblock In \emph{Proceedings of the IEEE Conference on Computer Vision and
  Pattern Recognition}, 8061--8069.

\bibitem[{Simonyan and Zisserman(2014)}]{simonyan2014very}
Simonyan, K.; and Zisserman, A. 2014.
\newblock Very deep convolutional networks for large-scale image recognition.
\newblock \emph{arXiv preprint arXiv:1409.1556}.

\bibitem[{Wang et~al.(2018)Wang, Wang, Zhou, Ji, Gong, Zhou, Li, and
  Liu}]{wang2018cosface}
Wang, H.; Wang, Y.; Zhou, Z.; Ji, X.; Gong, D.; Zhou, J.; Li, Z.; and Liu, W.
  2018.
\newblock Cosface: Large margin cosine loss for deep face recognition.
\newblock In \emph{Proceedings of the IEEE Conference on Computer Vision and
  Pattern Recognition}, 5265--5274.

\bibitem[{Wang et~al.(2019)Wang, Bao, Sun, Zhu, Cao, and
  Philip}]{wang2019private}
Wang, J.; Bao, W.; Sun, L.; Zhu, X.; Cao, B.; and Philip, S.~Y. 2019.
\newblock Private model compression via knowledge distillation.
\newblock In \emph{Proceedings of the AAAI Conference on Artificial
  Intelligence}, volume~33, 1190--1197.

\bibitem[{Wang et~al.(2021)Wang, Chen, Zhu, Chu, Tai, Wang, Li, Wu, Huang, and
  Ji}]{wang2021hififace}
Wang, Y.; Chen, X.; Zhu, J.; Chu, W.; Tai, Y.; Wang, C.; Li, J.; Wu, Y.; Huang,
  F.; and Ji, R. 2021.
\newblock HifiFace: 3D Shape and Semantic Prior Guided High Fidelity Face
  Swapping.
\newblock \emph{arXiv preprint arXiv:2106.09965}.

\bibitem[{Xu et~al.(2021)Xu, Yu, Hong, Zhu, Han, Liu, Ding, and
  Bai}]{xu2021facecontroller}
Xu, Z.; Yu, X.; Hong, Z.; Zhu, Z.; Han, J.; Liu, J.; Ding, E.; and Bai, X.
  2021.
\newblock FaceController: Controllable Attribute Editing for Face in the Wild.
\newblock In \emph{Proceedings of the AAAI Conference on Artificial
  Intelligence}, volume~35, 3083--3091.

\bibitem[{Yim et~al.(2017)Yim, Joo, Bae, and Kim}]{yim2017gift}
Yim, J.; Joo, D.; Bae, J.; and Kim, J. 2017.
\newblock A gift from knowledge distillation: Fast optimization, network
  minimization and transfer learning.
\newblock In \emph{Proceedings of the IEEE Conference on Computer Vision and
  Pattern Recognition}, 4133--4141.

\bibitem[{Yu et~al.(2018)Yu, Wang, Peng, Gao, Yu, and Sang}]{yu2018bisenet}
Yu, C.; Wang, J.; Peng, C.; Gao, C.; Yu, G.; and Sang, N. 2018.
\newblock Bisenet: Bilateral segmentation network for real-time semantic
  segmentation.
\newblock In \emph{Proceedings of the European conference on computer vision},
  325--341.

\end{thebibliography}
}

\end{document}